\documentclass[twoside,journey]{IEEEtran}
\usepackage{makecell}
\usepackage{hyperref}
\usepackage{array}
\usepackage{graphicx,amssymb,amsmath}
\usepackage{multicol}
\usepackage[noadjust]{cite}
\usepackage{setspace}
\usepackage{subcaption}
\usepackage{graphicx}
\usepackage{float}
\usepackage {url}
\usepackage{stfloats}
\usepackage{amsthm,pifont}
\usepackage{cases,subeqnarray}
\usepackage{bm,multirow,bigstrut}
\usepackage{amsmath, amsthm, amssymb}
\usepackage{textcomp}
\usepackage{latexsym,bm}
\usepackage{booktabs}
\usepackage{xcolor}
\usepackage{mathtools}
\usepackage{dsfont}
\usepackage{extarrows}
\usepackage{epsfig}
\usepackage{epsfig}
\usepackage{epstopdf}
\usepackage[noend]{algpseudocode}
\usepackage{algorithmicx,algorithm}
\usepackage{makecell}
\usepackage{colortbl}
\usepackage{booktabs}
\usepackage{graphicx} 
\usepackage{array}    
\usepackage{hyperref} 
\usepackage{booktabs}
\usepackage{enumitem}
\usepackage{diagbox}
\usepackage{tikz}
\usetikzlibrary{trees,shadows.blur}
\usepackage{forest}
\theoremstyle{plain}

\theoremstyle{plain}

\usepackage{amsmath}

\newcommand{\ignore}[1]{{{\color{yellow} }}}
\definecolor{blue-green}{rgb}{0.0, 0.87, 0.87}
\IEEEoverridecommandlockouts
\begin{document}

\title{From Agentification to Self-Evolving Agentic AI for Wireless Networks: Concepts, Approaches, and Future Research Directions}
\author{Changyuan Zhao, Ruichen Zhang, Jiacheng Wang, Dusit Niyato,~\IEEEmembership{Fellow,~IEEE}, Geng Sun, \\
Xianbin Wang,~\IEEEmembership{Fellow,~IEEE}, Shiwen Mao,~\IEEEmembership{Fellow,~IEEE}, and Abbas Jamalipour,~\IEEEmembership{Fellow,~IEEE}
\thanks{C. Zhao is with the College of Computing and Data Science, Nanyang Technological University, Singapore, and CNRS@CREATE, 1 Create Way, 08-01 Create Tower, Singapore 138602 (e-mail: zhao0441@e.ntu.edu.sg).
}
\thanks{R. Zhang, J. Wang, and D. Niyato are with the College of Computing and Data Science, Nanyang Technological University, Singapore (e-mail: ruichen.zhang@ntu.edu.sg; jiacheng.wang@ntu.edu.sg; dniyato@ntu.edu.sg).}
\thanks{G. Sun is with College of Computer Science and Technology, Jilin University, China 130012, (e-mail: sungeng@jlu.edu.cn).}
\thanks{X. Wang is with the Department of Electrical and Computer Engineering, Western University, London, ON, N6A 5B9, Canada (e-mail: xianbin.wang@uwo.ca).}
\thanks{S. Mao is with the Department of Electrical and Computer Engineering,
Auburn University, Auburn, USA (e-mail: smao@ieee.org).}
\thanks{A. Jamalipour is with the School of Electrical and Computer Engineering,
University of Sydney, Australia (e-mail: a.jamalipour@ieee.org).}
}

\maketitle
\vspace{-1cm}

\begin{abstract}
Self-evolving agentic artificial intelligence (AI) offers a new paradigm for future wireless systems by enabling autonomous agents to continually adapt and improve without human intervention. Unlike static AI models, self-evolving agents embed an autonomous evolution cycle that updates models, tools, and workflows in response to environmental dynamics. This paper presents a comprehensive overview of self-evolving agentic AI, highlighting its layered architecture, life cycle, and key techniques, including tool intelligence, workflow optimization, self-reflection, and evolutionary learning. 
We further propose a multi-agent cooperative self-evolving agentic AI framework, where multiple large language models (LLMs) are assigned role-specialized prompts under the coordination of a supervisor agent. Through structured dialogue, iterative feedback, and systematic validation, the system autonomously executes the entire life cycle without human intervention. A case study on antenna evolution in low-altitude wireless networks (LAWNs) demonstrates how the framework autonomously upgrades fixed antenna optimization into movable antenna optimization. 
Experimental results show that the proposed self-evolving agentic AI autonomously improves beam gain and restores degraded performance by up to 52.02\%, consistently surpassing the fixed baseline with little to no human intervention and validating its adaptability and robustness for next-generation wireless intelligence.

\end{abstract}
\begin{IEEEkeywords}
Agentification, self-evolving agentic AI, large language models, low-altitude wireless networks, 
\end{IEEEkeywords}
\IEEEpeerreviewmaketitle

\section{Introduction}




The concept of the Gödel Machine, proposed by Jürgen Schmidhuber, envisions a self-referential artificial intelligence~(AI) capable of provably improving itself by rewriting its own code~\cite{zhang2025darwin}. 
If realized, such systems would not only react to their environment but also actively reshape themselves, transforming every failure into a stepping stone toward optimal performance progressively. 
For years, however, this notion remained largely conceptual.

Building on this foundational idea, the paradigm of agentification, \textit{through the process of transforming static AI models into autonomous and adaptive agents}, has recently gained traction~\cite{zhang2025toward}.
This development underpins the emergence of self-evolving agentic AI~\cite{gao2025survey}, turning the theoretical promise of self-improvement into practical agent architectures.
Self-evolving agentic AI represents a new generation of autonomous agent systems that can continuously adapt and self-improve through dynamic interaction with their environments, effectively bridging the powerful yet static capabilities of AI models with the continual adaptability required by edge systems~\cite{fang2025comprehensive}. 
In practice, such agents leverage a synergy of learning techniques: reinforcement learning for trial-and-error optimization, self-supervised learning for extracting structure from raw data, curriculum learning for staged skill growth, and self-reflection for diagnosing weaknesses~\cite{gao2025survey}.




Although existing techniques such as continuous learning, life-long learning, incremental learning, and domain adaptation offer mechanisms to update static models, they typically rely on human intervention and only target isolated stages of the model life cycle~\cite{gao2025survey}. 
In contrast, self-evolving agents can autonomously execute the full AI agent life cycle without human intervention, from acquiring and curating new experiences to refining and integrating them to updating models and deploying the improved capabilities. This evolution cycle allows the agent to continuously enhance individual components such as the model, memory, prompts, tools, and workflow strategies efficiently and effectively. In doing so, the system dynamically adapts to changing tasks, dynamic contexts, and varying resources while ensuring safety, stability, and sustained performance.

Given the advantages of self-evolving agents, this emerging concept is increasingly becoming a practical reality. A representative example is Google DeepMind’s AlphaEvolve, an evolutionary coding agent that embodies the self-evolving paradigm in action\footnote{https://deepmind.google/discover/blog/alphaevolve-a-gemini-powered-coding-agent-for-designing-advanced-algorithms/}. AlphaEvolve iteratively performs closed-loop cycles of code generation, automated evaluation, and intelligent mutation, guided by large language models (LLMs) and evolutionary selection mechanisms. The system has achieved tangible impact by discovering algorithms surpassing 50-year-old benchmarks, reclaiming 0.7\% of Google’s global compute resources, and accelerating AI training.

With the emergence of 6G and next-generation communication technologies, edge and Internet of Things (IoT) devices are anticipated to operate with unprecedented levels of autonomy and intelligence~\cite{zhao2025edge}. 
Agentification provides an important step in this transition, as it transforms static AI models into autonomous agents capable of perception, reasoning, and action~\cite{jiang2025large}.
However, despite this progress, conventional AI agents remain predominantly static, require manual intervention for updates, and lack adaptability to the highly dynamic and heterogeneous wireless environments introduced by 6G.
In wireless systems, self-evolving agentic AI addresses dynamic environments by processing multimodal data, including signals, mobility, and network states, for decision-making purposes with minimal human intervention. Model selection is task-aware, aligning with objectives such as beam alignment and interference mitigation, e.g., using CNNs for spatial features and RNNs for temporal dynamics~\cite{zhao2025edge}. Lightweight and adaptive models support edge deployment and scalable updates. A UAV base station, for instance, can evolve with changing weather, new sensing modalities, or movable antennas, thereby improving overall system performance and adapting to the heterogeneous demands of 6G.


Therefore, this paper presents a comprehensive perspective on self-evolving agentic AI tailored for wireless systems. 
\textit{To the best of our knowledge}, this is the first work to explore the self-evolving agentic AI for intelligent wireless systems.
The key contributions of this work are summarized as follows:
\begin{itemize}
    \item We present a comprehensive overview of self-evolving agentic AI for wireless systems. We introduce its layered architecture, life cycle, and enabling techniques such as tool intelligence, workflow optimization, self-reflection, contextual adaptation, and evolutionary learning.

    \item We propose a multi-agent cooperative self-evolving framework, where multiple LLMs are assigned role-specialized prompts under the coordination of a supervisor agent. Through structured dialogue, iterative feedback, and systematic validation, the system autonomously executes the entire life cycle without human intervention.

    \item We conduct a case study on antenna evolution in low-altitude wireless networks (LAWNs), demonstrating how the proposed framework upgrades fixed antenna optimization into movable antenna optimization. The results show that our collaborative agentic AI improves the beam gain and achieves performance recovery of up to 52.02\% after degradation, validating its adaptability and robustness with minimal human involvement in model reformulation and upgrading.
    

 
\end{itemize}

We envision that this work can fuel ongoing and emerging research in self-evolving agentic AI for wireless communications. With this paper, researchers will be able to
\begin{itemize}
    \item  Gain a clear understanding of self-evolving agentic AI and its role in advancing intelligent wireless systems.  
    
    \item  Learn how to design and deploy proactive, adaptive wireless systems with self-evolving agentic AI.  
    
    \item 
    Incorporate self-evolving techniques, including tool intelligence, workflow optimization, self-reflection, and evolutionary learning, into wireless systems. 
\end{itemize}

\section{Overview of Agentic AI}

\begin{figure*}[htp]
    \centering
    \includegraphics[width= 0.80\linewidth]{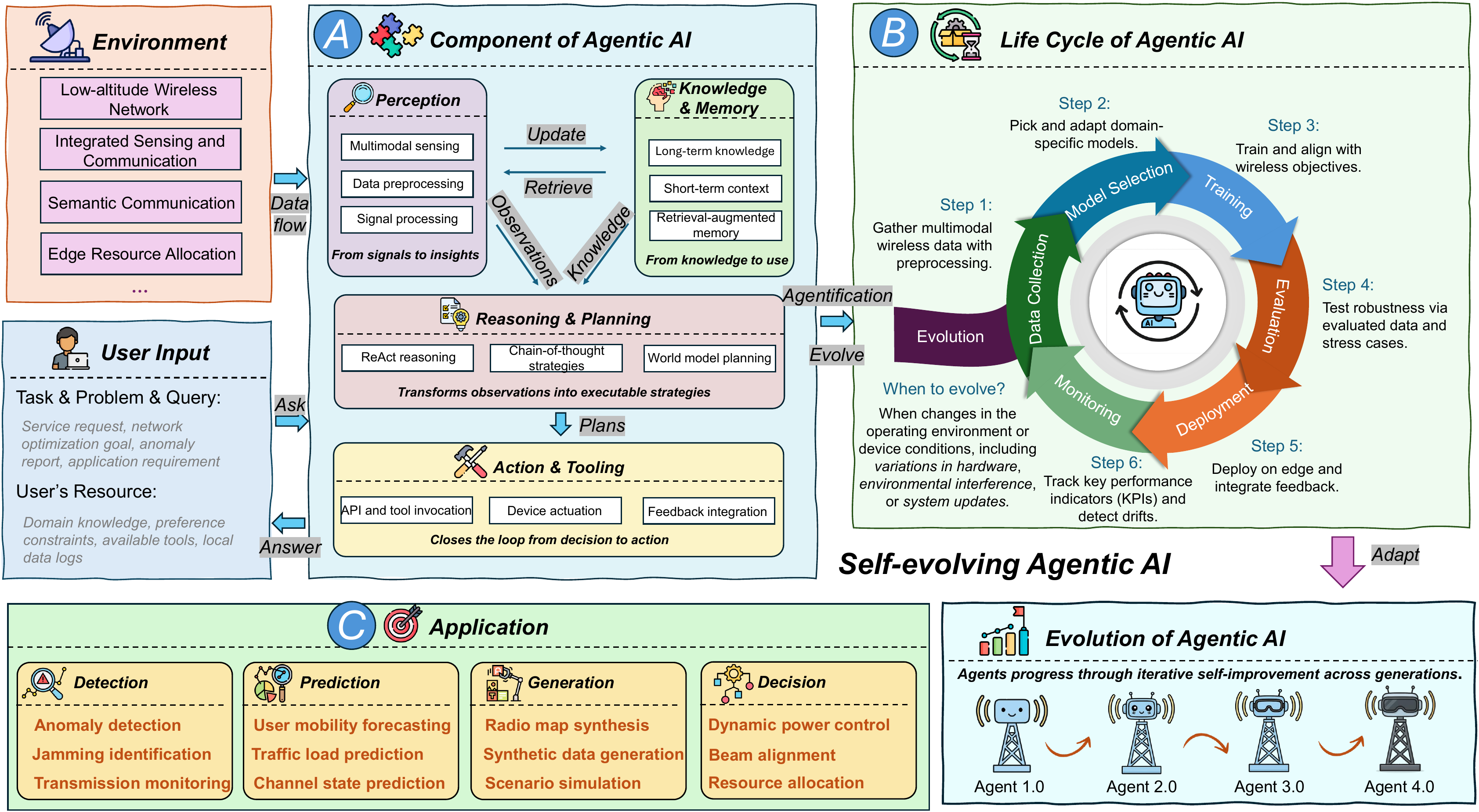}
    \caption{Illustration of self-evolving agentic AI for intelligent wireless networks.
Part A highlights the core components of agentic AI, including perception, knowledge and memory, reasoning and planning, and action and tooling.
Part B illustrates the life cycle of agentic AI, showing how self-evolution is triggered when key performance indicators degrade or conditions shift.
Part C presents the applications and evolution of agentic AI, where agents progress through iterative self-improvement to enhance detection, prediction, generation, and decision-making in wireless environments.}
    \label{fig:world}
\end{figure*}


Agentic AI refers to systems endowed with autonomous decision-making, adaptive learning, and proactive interaction capabilities, enabling them to achieve complex goals in dynamic and uncertain environments without constant human oversight~\cite{acharya2025agentic}. Unlike conventional AI models that passively respond to inputs, agentic AI systems actively perceive their environment, reason about actions, plan strategies, and interact with external tools to accomplish tasks.

Building on this concept, the process of agentification involves transforming traditional AI pipelines into such autonomous agents~\cite{zhang2025toward}. In wireless networks, this entails embedding AI modules with the ability to autonomously select and operate tools, adapt to evolving communication scenarios, and minimize reliance on manual interventions.

\subsection{Layered Architecture and Functional Roles of Agentic AI}

An agentic AI system can be viewed as a layered stack, with each layer providing specific functions and interacting seamlessly with the others~\cite{zhao2025edge}.
In current practice, LLMs serve as the core reasoning engine, orchestrating perception, memory, planning, and action through structured prompts and tool calls:

\begin{itemize}
\item \textbf{Perception Layer:}
The system gathers multimodal data from edge devices, IoT sensors, and RF frontends. Local models handle images, signals, or logs, with preprocessed outputs passed to LLMs for unified understanding and normalization.

\item \textbf{Knowledge and Memory Layer:}  
Through memory buffers, vector databases, and retrieval-augmented generation (RAG), agentic AI retrieves both long-term domain knowledge and short-term context, enabling edge-aware, privacy-preserving cognition.

\item \textbf{Reasoning and Planning Layer:}  
LLMs implement reasoning–action (ReAct) prompting, chain-of-thought reasoning, and predictive planning using world models to guide tasks such as~\cite{jiang2025large,zhao2025edge}:
\begin{itemize}
    \item \emph{Detection} of anomalies or spectrum misuse by fusing logs, signals, and sensor data in real time.
    \item \emph{Prediction} of mobility or channel conditions for proactive handovers or beamforming.
    \item \emph{Generation} of synthetic radio maps or user patterns in unseen environments.
    \item \emph{Decision-making} for tasks such as power control or beam alignment under uncertainty.
\end{itemize}

\item \textbf{Action and Tooling Layer:}  
Plans are executed through application programming interface (API) calls, control interfaces, or physical device actuation. LLMs manage tool use and refine strategies by incorporating execution feedback into subsequent loops.

\end{itemize}




This layered architecture equips agentic AI with essential capabilities: perception fuses multimodal inputs for situational awareness, knowledge and memory preserve long-term expertise and short-term context, reasoning and planning leverage LLMs to optimize decisions, and action and tooling execute strategies through APIs or device controllers~\cite{gao2025survey}. As agentic AI evolves, these layers must be reconfigured to sustain performance. For instance, a UAV base station upgraded with onboard cameras requires enhanced visual processing in the perception layer, while movable antennas demand updated knowledge representations to capture orientation- and mobility-dependent channel characteristics.

\subsection{Life Cycle of Agentic AI Development}

The progress toward agentification relies on a structured and iterative development life cycle.
Within intelligent wireless networks, this life cycle encompasses data collection, model design, training, evaluation, and operational feedback:

\subsubsection{Data Collection}
Developers gather and prepare datasets essential for building and refining the agent. This includes historical logs, annotated interaction transcripts, tool execution traces, and synthetic or simulated data. In wireless network contexts, such data are often sourced from distributed gateways or digital twins, with preprocessing, filtering, and privacy-preserving aggregation to ensure quality and compliance for subsequent training and evaluation~\cite{zhao2025edge}.

\subsubsection{Model Selection}
Engineers choose an appropriate foundation LLM or hybrid architecture that combines reasoning engines with task-specific models. Adaptation to the target domain is achieved through parameter-efficient fine-tuning, multimodal fusion, and integration with retrieval-augmented generation pipelines, enabling the agent to embed domain-specific knowledge and remain responsive to wireless environments~\cite{zhang2025toward}.

\subsubsection{Training}
Researchers establish the agent’s core capabilities through large-scale pretraining or domain-specific supervised learning. Alignment techniques such as reinforcement learning with human feedback (RLHF), RL from AI Feedback (RLAIF), and direct preference optimization~(DPO) are then applied to refine the model’s behavior toward objectives such as fairness, safety, and efficiency under the constraints of wireless systems~\cite{jiang2025large}.

\subsubsection{Evaluation}
Engineers assess the agent’s reasoning, planning, and decision capabilities using benchmark datasets, digital twin simulations, and pilot deployments. Evaluation should also include stress testing with adversarial prompts, noisy sensor inputs, and network degradation scenarios that reflect realistic wireless conditions, ensuring resilience and robustness before large-scale rollout.

\subsubsection{Deployment}

The system architect is responsible for packaging the agent for execution on edge infrastructure or IoT gateways, with careful attention to resource efficiency, scalability, and reproducibility. Deployment should be co-optimized with resource allocation, accounting for hardware heterogeneity, network constraints, and dynamic topologies. Operational feedback, such as failures or new tool integrations, should be continuously fed back into the development loop for ongoing adaptation and refinement~\cite{zhang2025toward}.

\subsubsection{Monitoring}
The agent continuously tracks performance, detects drifts or anomalies, and adapts to evolving spectrum conditions, user mobility, and interference patterns. Monitoring provides critical operational feedback that drives iterative improvement, ensuring sustained adaptability, robustness, and reliability throughout the lifecycle of the system in dynamic wireless environments~\cite{zhao2025generative}.

\subsection{Necessity of Self-Evolving in Agentic AI}

After agentification, most AI agents remain static, with knowledge and skills constrained by past training data, initial programming, or pre-configured environments. For example, GPT-4 is trained only up to December 2023\footnote{https://learn.microsoft.com/en-us/azure/ai-foundry/openai/overview}. In wireless communications, this means the agent may answer questions about 3GPP Releases 17 or 18 but fail to capture newer progress such as Release 19, which marks the early stage of 6G\footnote{https://www.3gpp.org/technologies/ran-rel-19}. While monitoring in the agentic AI life cycle can detect model drift, degraded accuracy, or new anomalies, updates still depend on human intervention, slowing adaptation and risking prolonged inefficiency.

Self-evolving is essential once monitoring reveals performance degradation. It can be viewed as automatically repeating the process of agentification in response to new requirements or environments, a “re-agentification” process. For example, when a UAV is equipped with a movable antenna, the agent must reconfigure its knowledge and decision modules to adapt to orientation-dependent channel dynamics.

\section{Self-Evolving Agentic AI}


In this section, we introduce the concept and enabling techniques of self-evolving agentic AI, and illustrate how it ensure continual adaptation and robust performance.

\subsection{Concept of Self-evolving Agentic AI}

Self-evolving agentic AI extends agentic AI by embedding an evolution cycle, comprising experience acquisition, refinement, model and tool updates, and redeployment, and is executed autonomously without human intervention~\cite{fang2025comprehensive}. Beyond carrying out tasks, these systems continuously improve their internal models, memory, prompts, tools, and workflow strategies in response to changing environments, task demands, and resource constraints. This self-directed growth enables long-term performance enhancement, resilience, and adaptability, making self-evolving agentic AI particularly valuable for mission-critical wireless applications facing persistent uncertainty and change.

\subsection{Self-evolving Techniques of Agentic AI}

This subsection outlines five core self-evolving techniques that enable agents to continuously improve their performance. Each technique contributes to enhanced robustness, efficiency, and intelligence in wireless communication systems..

\subsubsection{Tool Intelligence}


Self-evolving agents enhance their capabilities by generating and refining tools such as APIs, scripts, or functions. Methods like Toolformer, ToolLLM, and ToolGen show how agents can autonomously design and adapt toolsets~\cite{schick2023toolformer}. In wireless networks, this enables rapid integration of new signal processing or optimization modules, reducing reliance on pre-engineered solutions and supporting agile adaptation to dynamic conditions.

\subsubsection{Workflow Optimization}


Workflow evolution enables autonomous optimization of task pipelines and system architectures. Agents can restructure operations, integrate heterogeneous models, or coordinate teams of agents, as shown by AFlow, ADAS, and EvoFlow~\cite{zhang2025evoflow}. This capability improves wireless resource allocation and large-scale coordination, enhancing overall performance and resilience.

\subsubsection{Self-Reflection}

Self-reflection enables agents to evaluate their own actions, critique failures, and refine strategies without external supervision. Techniques such as Reflexion, Self-Refine, and AdaPlanner illustrate this process, where agents iteratively analyze mistakes and revise future plans accordingly~\cite{sun2023adaplanner}.
By integrating self-critique, agents can detect ineffective decisions and autonomously adjust strategies, enhancing robustness in volatile environments.

\subsubsection{Contextual Adaptation}


Adaptation involves memory evolution, prompt optimization, and AutoML to flexibly update context and internal models~\cite{fang2025comprehensive}. Memory evolution supports storing, forgetting, and reorganizing knowledge for coherence, while prompt optimization dynamically refines instructions without retraining. AutoML further searches for efficient architectures under resource limits. In wireless networks, these mechanisms help agents preserve knowledge of mobility, signal statistics, and traffic patterns, enabling faster task adaptation and lightweight deployment on edge devices.

\subsubsection{Evolutionary Learning}


Learning-driven evolution leverages rewards, demonstrations, and population-based methods for continual improvement. Agents optimize policies via throughput or latency rewards, refine behaviors from demonstrations, and evolve collectively using strategies such as RAGEN, DYSTIL, and STaR~\cite{gao2025survey}. Through natural feedback and expert demonstrations, they can refine policies for dynamic multi-user, multi-cell, and drone-assisted networks.

\subsection{Self-evolution in Agentic AI Layers}



Self-evolving agentic AI enables continuous improvement across different functional layers. Each round of evolution may involve completing all or part of the AI life cycle, depending on system requirements. Beyond degraded performance, evolution can also be triggered by the availability of new datasets, novel model architectures, hardware upgrades, or shifts in operating environments. In wireless networks, self-evolution manifests differently across perception, knowledge, reasoning, and action layers.

\subsubsection{Perception Layer}
At the perception layer, evolution is often triggered by hardware upgrades or new sensing modalities. For example, a UAV may be equipped with additional cameras, or a fixed antenna array may be upgraded to a movable one. Such changes require the perception outputs to adapt accordingly~\cite{xu2025enhancing}. By leveraging workflow optimization, the agent can seamlessly integrate new hardware and extend its sensing capabilities without manual redesign.

\subsubsection{Knowledge and Memory Layer}
Evolution at the knowledge layer is typically driven by updates in protocols, standards, or environments. For instance, the introduction of new 3GPP releases or the deployment of UAVs in previously unseen environments may render the old knowledge base insufficient~\cite{3gpp.36.331}. Here, tool intelligence and contextual adaptation enable the agent to search, filter, and restructure relevant knowledge, dynamically building new models or databases that reflect the latest context.

\subsubsection{Reasoning and Planning Layer}
Changes in hardware or environment also necessitate updates in reasoning and planning. For example, when upgrading from fixed to movable antennas, the agent must expand its reasoning beyond simple beam direction to also consider antenna positions~\cite{10906511}. This demands workflow optimization to redesign the decision pipeline and self-reflection to refine reasoning strategies, ensuring robust planning in evolving conditions.

\subsubsection{Action and Tooling Layer}
At the action and tooling layer, evolution refers to the agent’s ability to perform new actions or develop new tools. For instance, integrating image data from newly mounted UAV cameras requires the generation of novel processing pipelines~\cite{xu2025enhancing}. Evolutionary learning enables the system to explore and establish suitable new actions, while tool intelligence allows the creation of specialized tools that enhance the agent’s effectiveness in dynamic tasks.

\begin{figure*}[htp]
    \centering
    \includegraphics[width= 0.95\linewidth]{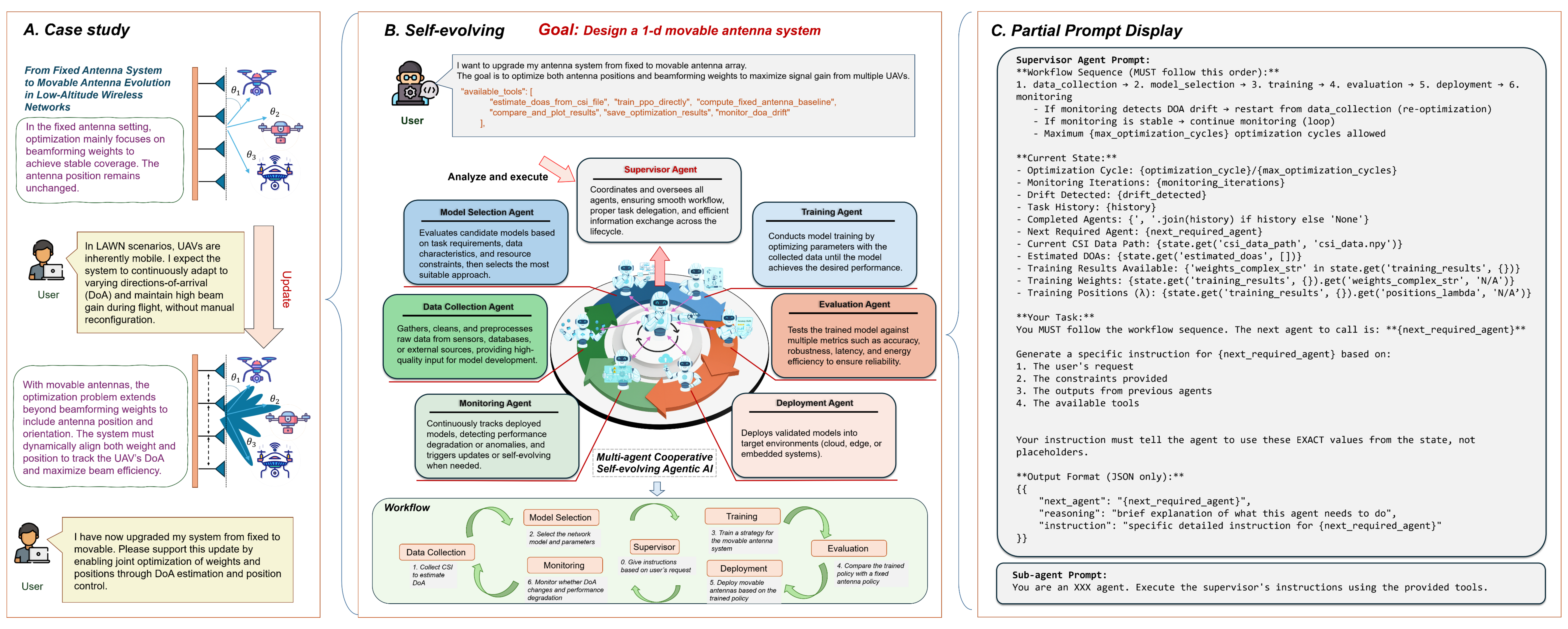}
    \caption{
    Multi-agent cooperative self-evolving agentic AI framework for intelligent wireless networks.
Part A illustrates a case study transitioning from fixed antenna optimization to movable antenna evolution in LAWNs. Part B presents the self-evolving pipeline, where multiple AI agents are orchestrated by a supervisor agent to support dynamic antenna evolution. Part C demonstrates a partial prompt example, showing how the supervisor agent leverages structured state variables and user instructions to guide the sequence of sub-agents.
    }
    \label{fig:frame}
\end{figure*}

\section{Case Study: From Fixed Antenna to Movable Antenna Evolution in LAWN Communication}

To bridge the general concepts with practical implementation, we illustrate how the core self-evolving capability of agentic AI can be instantiated in a wireless scenario. Specifically, the case study demonstrates how tool intelligence, workflow optimization, and self-reflection transform a fixed antenna system into a movable antenna system in LAWNs. By doing so, we present the entire evolving process, showing how self-evolution can be applied in real-world wireless deployments.

\subsection{System Model}

We consider a base-station antenna system in LAWNs that simultaneously serves multiple UAVs. We model the scenario as a multi-user downlink communication system from the base station to UAVs, in which UAVs are located at different directions of arrival (DoAs), measured relative to the axis of the antenna array of the base station.

As an initial version, the base station is equipped with a fixed linear antenna array. The antenna elements are uniformly spaced, and beamforming is performed with predetermined steering directions. This setup offers reliable performance when UAV locations are static or slowly varying. 
To enhance system performance, the users then extend the system to a movable antenna array~\cite{10906511}, where each array element can be repositioned along a one-dimensional line, as shown in Fig. \ref{fig:frame} Part A. The design objective is to jointly determine the antenna positions and beamforming weights to maximize the overall received gain for all UAV users.

The system operates at a carrier frequency of $2.4~\text{GHz}$, corresponding to a wavelength of about $\lambda = 0.125~\text{m}$. The array is composed of eight movable antennas, each constrained by a minimum spacing of half a wavelength to avoid mutual coupling, while the movement of the array elements is restricted to a symmetric range of plus or minus five wavelengths. 
The UAVs are assumed to vary continuously over $T$ steps,
and the system adapts its antenna configuration accordingly to enhance service quality.

As shown in Fig.~\ref{fig:frame}, the system leverages channel state information (CSI) to estimate the DoAs of UAVs, and subsequently optimizes the antenna configuration based on these estimates. In Fig.~\ref{fig:frame} Part B, the system upgrades the antenna system from a fixed to a movable setup and provides additional tools that can be invoked by the framework. Building on these inputs, the self-evolving agentic AI framework adapts the software accordingly, enabling the system to autonomously reconfigure itself for movable antenna optimization. This design ensures that the network can flexibly accommodate dynamic UAV deployments in LAWNs while reducing engineering effort.


\subsection{Multi-agent Cooperative Self-evolving Agentic AI}


Typically, upgrading from fixed antennas to movable antennas requires hardware updates and software reconfiguration, which often involves human intervention. In this subsection, we present a multi-agent collaborative agentic AI framework for antenna self-evolution, where each stage of the AI life cycle is assigned to a specialized agent, and a supervisor agent orchestrates the overall workflow~\cite{qian2023chatdev}. This design enables the entire pipeline, from data collection to monitoring, to autonomously coordinate antenna self-evolution in a coordinated and adaptive manner with minimal human intervention or supervision.

\subsubsection{Multi-agent Collaboration}

In the collaborative framework, multiple LLMs are assigned distinct roles by well-designed prompts to emulate organizational workflows, as shown in Fig.~\ref{fig:frame} Part C. Each role focuses on a single responsibility, thereby reducing overlap and ensuring precise contributions. 

At the center of this framework is the \textbf{Supervisor Agent}, which coordinates role-specific agents, manages state variables, and integrates feedback. It can also perceive the environment via tools, information, or human commands to detect system changes, such as the shift from fixed to movable antennas, and initiate a self-evolution cycle. By invoking agents according to current conditions and outcomes, the supervisor agent enables the system to seamlessly self-evolve with minimal human supervision. 

Therefore, specialized agents contribute their domain expertise and refine outputs through structured coordination under the supervisor agent. \textbf{Tool intelligence} is achieved as agents with designated roles autonomously generate new optimization utilities and iteratively refine existing ones through feedback exchanges. \textbf{Workflow optimization} emerges from the supervisor agent's ability to dynamically organize the sequence of agents across the life cycle. \textbf{Self-reflection} is achieved through monitoring and evaluation agents that continuously assess intermediate results, identify mismatches or errors, and trigger corrective updates. These capabilities implement the principles of self-evolving agentic AI in real-world scenarios.

\subsubsection{Planning and Feedback for Self-Evolution}

The collaborative framework transforms the traditionally human-driven AI life cycle into a fully autonomous process. Through structured communication and feedback, the agents collectively achieve self-evolution. 

For example, during \textbf{data collection}, the Data Collection Agent processes CSI to estimate UAV DoAs once the antenna system evolves from fixed to movable. In \textbf{model selection}, the Model Selection Agent evaluates candidate optimization approaches and selects suitable neural networks for handling joint weight and position optimization. During \textbf{training}, the Training Agent develops models to optimize beamforming weights and antenna positions based on the selected method. In the \textbf{evaluation} stage, the Evaluation Agent verifies performance across metrics such as beam gain, robustness, and efficiency, providing structured feedback. This feedback loop demonstrates \textit{self-evolution}, where the system repeatedly revises and improves its models until requirements are met. For \textbf{deployment}, the Deployment Agent packages the optimized models for execution in realistic environments, ensuring reproducibility and compatibility. Finally, in \textbf{monitoring}, the Monitoring Agent continuously tracks performance, detects DoA drifts or degradations, and triggers re-optimization by instructing the supervisor agent to restart the cycle when necessary.

\begin{figure}[htp]
    \centering
    \includegraphics[width= 0.73\linewidth]{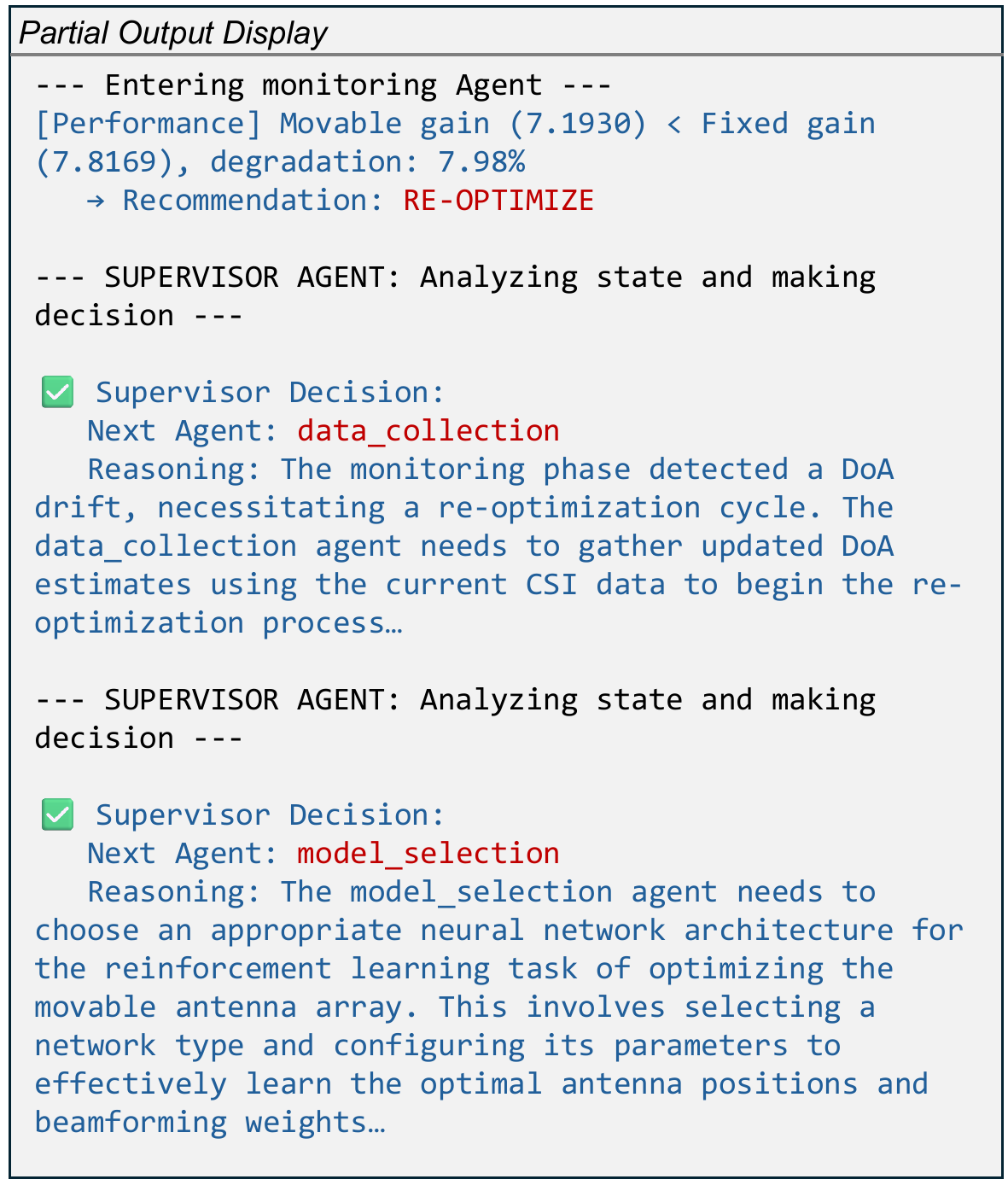}
    \caption{Collaborative interaction of LLM-driven agents. The monitoring agent detects performance degradation, and the supervisor invokes data collection and model selection agents, showing autonomous evolution without human intervention.}
    \label{fig:ex1}
\end{figure}

In Fig.~\ref{fig:ex1}, we present the decision-making results of the Supervisor Agent, which analyzes monitoring feedback and determines the next agent to invoke for re-optimization. More detailed prompts and implementation codes can be found in our open-source repository\footnote{https://github.com/ChangyuanZhao/LAE\_evolving}.

\subsection{Numerical Results}

\subsubsection{Experimental Setup}

Our experiments are conducted on a Linux server running Ubuntu 22.04 and equipped with an NVIDIA RTX A6000 GPU. The collaborative LLMs used are GPT-4o\footnote {https://openai.com/index/openai-api/}.
integrated and orchestrated through the LangGraph framework\footnote{https://www.langchain.com/langgraph}.

We first construct the baseline system using a fixed antenna array serving 3 UAVs, where only the beamforming weights can be optimized while the antenna element positions remain unchanged. This fixed configuration serves as both the initial benchmark for comparison and the reference baseline during the evaluation and monitoring of the movable antenna results.
Building on this baseline, the system is then upgraded to a movable antenna setting, in which both antenna positions and beamforming weights are adaptively optimized under the coordination of the collaborative agentic AI framework.


\subsubsection{Performance Analysis}

\begin{figure}[htp]
    \centering
    \includegraphics[width= 0.65\linewidth]{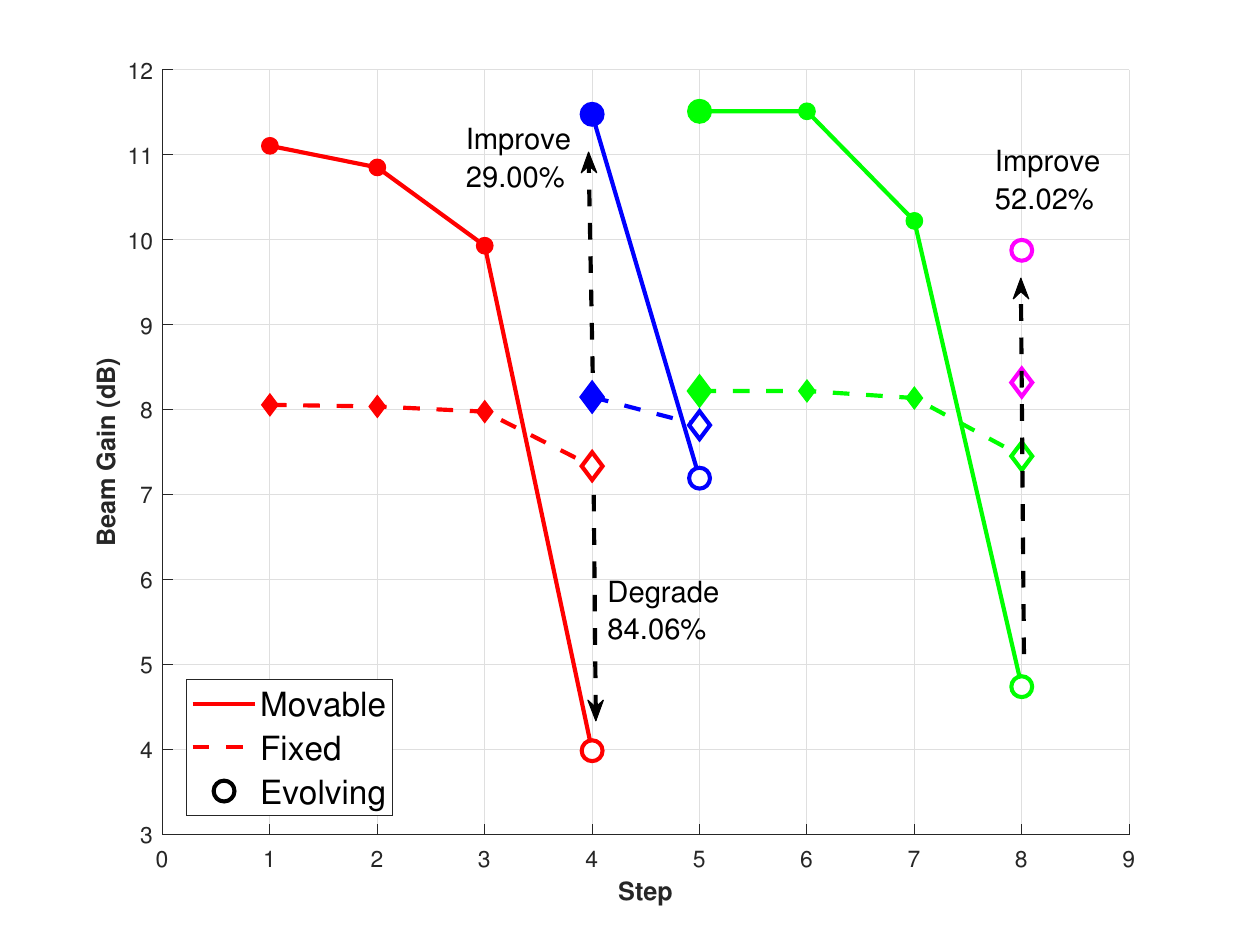}
    \caption{Comparison of sum beam gain. Solid lines denote movable antennas, dashed lines denote fixed antennas, and hollow circles mark evolution.
    }
    \label{fig:ex2}
\end{figure}

As shown in Fig.~\ref{fig:ex2}, our system first evolves from the fixed antenna baseline to the movable setting, improving the beam gain from 8.056~dB to 11.105~dB. After this initial upgrade, the framework enters the monitoring phase, where it continuously evaluates UAV angular variations and compares the current movable strategy against the fixed baseline to decide whether further evolution is required. 



As shown in Fig.~\ref{fig:ex2}, our system first evolves from the fixed antenna baseline to the movable setting, improving the beam gain from 8.056~dB to 11.105~dB. After this upgrade, the framework enters the monitoring phase, where it continuously evaluates UAV angular variations and compares the current movable strategy against the fixed baseline to decide whether further evolution is required. 

For example, when the movable gain drops to 3.9847~dB, which is 84.06\% lower than the fixed baseline of 7.8169~dB, the framework triggers re-optimization and recovers the performance to 11.105~dB, corresponding to a 29.00\% improvement. Another self-evolution further achieves a 52.02\% gain over a degraded value. These results demonstrate that the system can automatically adapt to environmental changes, restore lost performance, and consistently maintain superiority over the fixed baseline.

\section{Future Directions}

\subsubsection{Security, Safety, and Objective Alignment}


As self-evolving agentic AI updates its code, models, and tools, new security and safety risks arise. Threats include malicious code, poisoned data, and unverified tools, while over-optimization may favor throughput at the expense of latency, fairness, or energy efficiency. In drone networks, strategies may expand coverage but quickly drain batteries, and near-field communications amplify risks of adversarial manipulation or misaligned objectives. Future work should emphasize robust agent evolution protocols, tool verification, and multi-objective optimization to ensure secure and aligned adaptation.

\subsubsection{Tool Interoperability Across Ecosystems}

Seamless integration of diverse tools remains a bottleneck for self-evolving agents. Many widely used platforms, such as Sionna or MATLAB-based simulators, lack standardized APIs for autonomous agent access. While manual code-level integration is feasible, the absence of native agent-level operability restricts autonomy and hinders flexible evolution across heterogeneous software stacks. Addressing this challenge requires the development of open APIs, cross-platform tool wrappers, and federated tool marketplaces, enabling self-evolving agents to dynamically integrate simulation, optimization, and signal processing utilities across far-field and near-field scenarios.


\section{Conclusion}





In this paper, we have introduced self-evolving agentic AI for intelligent wireless networks, emphasizing its layered architecture, autonomous life cycle, and key techniques such as tool evolution, workflow optimization, and self-reflection. We have proposed a multi-agent collaborative agentic AI framework that enables autonomous self-improvement, spanning from data collection to monitoring. A case study on antenna evolution in LAWNs demonstrated that the framework ensures the automatic detection and correction of performance degradations without human intervention or supervision, showing both reliability and quantitative gains in dynamic wireless environments.

\bibliography{Ref}

\end{document}